%% file: main.tex
\documentclass{article}

\usepackage[utf8]{inputenc}
\usepackage{siunitx}
\usepackage{graphicx}
\usepackage{amsmath}
\usepackage{booktabs}
\usepackage{array}
\usepackage{multirow}
\usepackage{tabularx}
\usepackage{xcolor}
\usepackage{tikz}
\usetikzlibrary{shapes.geometric, arrows.meta, positioning, fit, backgrounds}
\usepackage{pgfplots}
\pgfplotsset{compat=1.18}
\usepackage{caption}
\usepackage[hidelinks]{hyperref}
\usepackage{cite}
\usepackage{balance}   

\newcolumntype{L}[1]{>{\raggedright\arraybackslash}p{#1}}
\newcolumntype{C}[1]{>{\centering\arraybackslash}p{#1}}

\definecolor{safetyred}{RGB}{229,62,62}
\definecolor{compblue}{RGB}{49,130,206}
\definecolor{comfgreen}{RGB}{56,161,105}
\definecolor{effamber}{RGB}{214,158,46}
\definecolor{navyblue}{RGB}{30,39,97}

\newcommand{\propbadge}[2]{%
  \textcolor{#1}{\textbf{[#2]}}%
}

\title{\LARGE \bf
From Operational Design Domain to Action:\\[4pt]
A Systematic Behavioral Taxonomy for Autonomous Driving
}
\author{Chaitanya Shinde$^{1}$, Hadi Hajieghrary$^{2}$, Miguel Hurtado$^{3}$
\thanks{$^{1}$Chaitanya Shinde {\tt\small chaitanya.shinde@torc.ai}, 
$^{2}$Hadi Hajieghrary {\tt\small hadi.hajieghrary@torc.ai},
$^{3}$Miguel Hurtado {\tt\small miguel.hurtado@torc.ai}
}}

\date{}
\begin{document}
\maketitle

\input{sections/00_abstract}
\input{sections/01_introduction}

\input{sections/02_related_work}
\input{sections/03_framework}

\input{sections/04_taxonomy}

\input{sections/05_deepdives}

\input{sections/06_scenarios}

\input{sections/07_discussion}
\input{sections/08_conclusions}

\balance   
\bibliographystyle{ieeetr}
\bibliography{refs}

\end{document}

%% file: sections/00_abstract.tex
\section{Abstract}

Operational Design Domain (ODD) specifications describe \emph{where} an
automated driving system (ADS) is permitted to operate, but they do not
prescribe \emph{what} the ADS must demonstrably do once deployed within
that domain. This gap between operating condition specification and
behavioral validation represents a critical unresolved challenge in ADS
safety assurance. This paper presents a structured, standards-grounded
taxonomy of 21 behavioral competencies organized across three operational
domains-Highway (HWY), Urban (URB), and Hub (HUB)-derived
systematically from the PEGASUS six-layer model-based ODD.
Each behavior is decomposed along longitudinal and lateral control axes
and characterized against a four-property framework:
\textbf{Safety} (gap maintenance, conflict avoidance, kinematic stability),
\textbf{Compliance} (legal rules and behavioral norms),
\textbf{Comfort} (rider dynamics and trust), and
\textbf{Efficiency} (mission completion and product-level metrics).
We further demonstrate that the crossing of ODD layer parameterizations
with behavioral competency specifications yields concrete scenario families
suitable for systematic behavioral testing and SOTIF coverage
evidence. The taxonomy is grounded in AVSC00008202111, SAE~J3237, and
SAE~J3016, and is validated as an operational specification layer
through its deployment in a rule-enforced trajectory optimization system.
The Hub domain is identified as a structurally distinct, underspecified
domain warranting dedicated research attention.

\vspace{6pt}
\noindent\small\textit{This paper presents an illustrative behavioral
taxonomy developed for research and engineering discussion purposes.
It reflects the views of the individual authors and was written based
on the state-of-the-art automated driving methodology understood at
the time of writing. It is not a complete or comprehensive work, and
its contents should be revisited as advances are made in technology,
standards, and applicable law. Nothing in this paper is intended to,
and nothing herein shall be construed to, establish, create, or
represent a minimum or maximum standard of care, a state-of-the-art
benchmark, or a complete specification of any product, system, or
safety validation process actually deployed by Torc Robotics, Inc.
Use of terms such as `safety,' `compliance,' or related terminology
refers to defined taxonomic categories used for engineering
classification purposes only and does not represent a claim, warranty,
or admission regarding the actual safety performance, compliance
status, or risk profile of any Torc product or system.}
\normalsize
\vspace{6pt}

%% file: sections/01_introduction.tex
\section{Introduction}

The deployment of automated driving systems (ADS) at commercial scale
requires a clear, testable answer to a deceptively simple question:
\emph{what must the ADS be able to do inside its operational design domain?}
Operational Design Domain (ODD) specifications-structured via frameworks
such as the PEGASUS six-layer model \cite{9400833}- define the
environmental and infrastructural conditions under which an ADS is designed
to function. They describe road geometry, traffic infrastructure, moving
object classes, environmental conditions, and digital information layers.
What they do not, in general, define is the behavioral competency expected of the ADS.

This gap has material consequences. Without a systematic enumeration of
behavioral competencies, safety validation efforts lack a complete coverage
target; behavioral compliance becomes a loosely defined concept varying
across teams and jurisdictions; and scenario-based testing lacks a
principled generation mechanism. The absence of such a layer is increasingly
recognized as a barrier to ADS certification and public trust 

This paper addresses the gap by presenting a \textbf{
behavioral taxonomy} for ADS, grounded in existing standards \cite{avsc2021} and structured
for practical use by safety, planning, and validation engineers. The
taxonomy organizes 21 behaviors across three operationally
distinct domains, provides a uniform anatomy for each behavior, and
demonstrates a formal mechanism for generating scenario families from the
crossing of ODD parameters and behavioral specifications.
Note: This behavior taxonomy is just for illustrative purposes and is one of the many ways one can define and group behaviors.

\subsection{Paper Objectives}

This paper pursues three objectives:

\begin{enumerate}
  \item \textbf{Derivation}: Demonstrate a systematic methodology for
    deriving operational domains and behaviors from ODD layer
    composition, using the PEGASUS model as the ODD substrate.
  \item \textbf{Taxonomy}: Present a structured, standards-grounded
    enumeration of 21 behaviors across Highway, Urban, and
    Hub domains/ use-cases, decomposed along longitudinal/lateral control axes and
    characterized via a four-property framework.
  \item \textbf{Operationalization}: Show that ODD parameter
    instantiation crossed with behavioral competency specifications yields
    concrete, testable scenario families aligned with ISO/PAS~21448
    (SOTIF) evidence generation.
\end{enumerate}

\subsection{Paper Organization}

Section~2 reviews related work and standards grounding.
Section~3 defines the behavioral framework including the
four-property characterization.
Section~4 presents the full 21-behavior taxonomy across three domains.
Section~5 provides deep-dive competency specifications for one behavior
per domain.
Section~6 formalizes the ODD $\times$ Behavior $\rightarrow$ Scenario
mapping.
Section~7 discusses the DDT/DDT-F dimension and the Hub domain gap.
Section~8 concludes with a summary of contributions and future directions.

%% file: sections/02_related_work.tex
\section{Background and Related Work}

\subsection{Standards Foundations}

The behavioral framing in this paper builds on four
standards-track documents.

\textbf{AVSC00008202111}~\cite{avsc2021} defines a behavioral competency
as a \emph{demonstrated capability in an ODD context with observable,
measurable outcomes}. It provides a framework for classifying ADS
behaviors and identifying evaluation criteria, and serves as the primary
source for the competency structure used in this taxonomy.

\textbf{Applicable Metrics}: Dynamic Driving Task Assessment
(DA) metrics from SAE J3237~\cite{j3237} applicable to this behavior,
organized by the J3237 taxonomy: Black Box metrics (computable without
ADS data), Grey Box metrics (requiring ADS status data), and White Box
metrics (requiring full ADS internal data).

\textbf{SAE J3016}~\cite{j3016} provides foundational vocabulary:
Dynamic Driving Task (DDT), DDT Fallback (DDT-F), Object and Event
Detection and Response (OEDR), and the ODD definition used throughout.

\textbf{ISO/PAS 21448 (SOTIF)}~\cite{sotif2022} defines the safety of
the intended functionality framework. The
scenario family generation mechanism in Section~6 is explicitly designed
to produce SOTIF-aligned coverage evidence.

\textbf{ISO 26262}~\cite{iso26262} addresses functional safety of
electrical and electronic systems and provides the systematic-fault
and random-hardware-fault assurance backbone against which SOTIF and
behavioral competency evidence are typically composed in a complete
safety case. \textbf{UL 4600}~\cite{ul4600} complements these with a
goal-based safety case standard specifically for autonomous products,
and \textbf{ISO/PAS 8800}~\cite{iso8800} extends this landscape to
safety-related properties of AI/ML elements within road vehicles,
addressing insufficiencies not covered by ISO 26262 or SOTIF alone.
The taxonomy in this paper is intended to sit alongside, not replace,
these standards: it supplies the behavioral competency layer that
such safety cases must reference as evidence.

\subsection{ODD Modeling}

The PEGASUS project~\cite{9400833} produced the most widely adopted
structured ODD model for ADS scenario-based testing: a six-layer
decomposition covering road geometry, road furniture and rules, temporary
modifications, moving objects, environmental conditions, and digital
information. This model is used directly in Section~3 to derive the
three operational domains and in Section~6 as the parameterization
substrate for scenario generation.

Subsequent work, including the ASAM OpenSCENARIO~\cite{openscenario}
and ASAM OpenDRIVE formats, operationalize portions of the PEGASUS
model in simulation toolchains. Our taxonomy is agnostic to simulation
format but maps onto these representations.

\subsection{Behavioral Taxonomy and Scenario Generation}

Existing behavioral taxonomies for ADS tend to be either (a) use-case
catalogs organized by traffic situation~\cite{ulbrich2015}, (b)
maneuver-level libraries for motion planning~\cite{hubmann2018}, or (c)
regulatory compliance checklists~\cite{nhtsa2022}. To our knowledge, existing work has not systematically derived behaviors from ODD layer composition, decomposes them along
longitudinal/lateral control axes, or characterizes them against a
unified property framework.

Scenario-based testing literature~\cite{menzel2018,riedmaier2020} has
addressed the challenge of scenario space coverage, typically operating
at the concrete or logical scenario level. Ontology-based approaches to
scene generation~\cite{bagschik2018} and keyword-based methods for
translating functional scenarios into executable logical
scenarios~\cite{menzel2019logical} provide complementary machinery for
instantiating the scenario families formalized in Section~6, but do not
themselves derive the behavioral competencies being tested. Our
contribution operates one
level above: at the behavioral competency level, providing the
intermediate layer between ODD specification and scenario instantiation.

Coverage adequacy for scenario-based validation remains an active
research area. Amersbach and Winner~\cite{amersbach2019} show a
persistent gap between statistically required and computationally
feasible scenario counts for highly automated vehicles, and situation-
and combinatorial-coverage criteria~\cite{alexander2015,tao2019} have
been proposed as tractable proxies. The Behavioral Competency Coverage
Score identified as future work in Section~\ref{sec:discussion}
is intended to provide a coverage criterion at the behavioral level,
complementary to these parameter- and situation-level approaches.

\subsection{Rule-Based Planning and Behavioral Compliance}

Recent work on rule-aware ADS planning~\cite{rector2026} has demonstrated
that explicit behavioral rule catalogs, organized by priority hierarchy,
can substantially reduce behavioral violations in closed-loop simulation.
The taxonomy presented in this paper provides the behavioral specification
layer from which such rule catalogs are derived, forming a complementary
pair: the taxonomy defines \emph{what} must be demonstrated; the
rule-enforcement layer defines \emph{how} it is achieved.

\subsection{Formal Safety Models and Safety Argumentation}

Formal models such as Responsibility-Sensitive Safety
(RSS)~\cite{rss2017} define mathematically verifiable rules for
longitudinal and lateral safe distances and have influenced the
kinematic stability sub-component of the Safety property
(Section~\ref{sec:4prop}). RSS operates at the maneuver/control level;
the taxonomy in this paper operates at the behavioral competency level
and can incorporate RSS-derived thresholds as one implementation choice
for the Safety Envelope metrics defined in J3237~\cite{j3237}.

Structuring the resulting evidence into a coherent safety case is a
separate concern from generating the evidence itself. Goal Structuring
Notation (GSN)~\cite{kelly2004gsn} provides a graphical safety argument
notation widely used to link claims, evidence, and context in ADS
safety cases, including in conjunction with standards such as UL
4600~\cite{ul4600}. The behavioral competency coverage evidence
generated via Equation~\ref{eq:scengen} is intended to serve as GSN-
compatible evidence nodes within such an argument structure, though
formalizing that mapping is left as future work
(Section~\ref{sec:future}).

Testing methodology challenges specific to ADS - non-deterministic
algorithms, inductive learning components, and driver-out-of-the-loop
operation - are cataloged by Koopman and
Wagner~\cite{koopman2016}, and the NHTSA testable-cases
framework~\cite{thorn2018} provides an early structured template for
ODD and OEDR-organized test case development that AVSC00008202111 and
this taxonomy both build upon.

%% file: sections/03_framework.tex
\section{Behavioral Framework}

\subsection{Defining a Behavioral Competency}

Following AVSC00008202111~\cite{avsc2021}, we define a \textbf{behavioral
competency} as a demonstrated capability of an ADS to execute a
goal-oriented action within a specified ODD context, with observable and
measurable outcomes against defined acceptance criteria. This definition
distinguishes three levels of behavioral description:

\begin{itemize}
  \item \textbf{Maneuver}: a physical control action (e.g., apply brake,
    steer left) - atomic, no goal semantics.
  \item \textbf{Behavior}: a goal-oriented action within the DDT
    (e.g., change lanes to avoid obstacle) - compositional, context-free.
  \item \textbf{Competency}: behavior + ODD context + evaluation criteria
    (e.g., execute lane change in rain at \SI{105}{\kilo\meter\per\hour}
with a 2\,s target gap, measured against J3237 DA metrics including
Safety Envelope Violation and Aggressive Acceleration Violation (AAV) thresholds~\cite{j3237})
    - the unit of specification and validation.
\end{itemize}

Competency is therefore the appropriate unit for taxonomy construction:
it captures \emph{what} the ADS must do, \emph{where} it must do it,
and \emph{how well} it must perform.

\subsection{The Four-Property Framework}
\label{sec:4prop}

Every behavioral competency is simultaneously governed by four
co-active properties. These are not a priority ordering; they are
concurrent constraints whose mutual tension defines where specification
is hardest.

\begin{description}
  \item[\propbadge{safetyred}{Safety}]
    Encompasses three distinct sub-components operating at different
    timescales: (1)~\emph{gap maintenance} - preventive, continuous
    longitudinal and lateral headway management (TTC, THW, minimum
    clearance); (2)~\emph{conflict and collision avoidance} - reactive
    OEDR response to dynamic hazards including cut-in vehicles, VRUs,
    and emergency vehicles; (3)~\emph{kinematic stability} - operation
    within the safe dynamic envelope defined by lateral/longitudinal
    acceleration limits, jerk bounds, and tire slip margins.
    A behavioral specification addressing only one sub-component is
    incomplete with respect to the full safety property.

  \item[\propbadge{compblue}{Compliance}]
    Encompasses two layers: (1)~\emph{legal compliance} - adherence to
    traffic laws, signal states, speed limits, right-of-way rules, and
    regulatory obligations; these are codified and largely testable
    against ground truth; (2)~\emph{behavioral norm compliance} -
    courtesy, predictability, and defensive posture toward other road
    users; these are socially expected but not always legally mandated,
    and harder to specify formally. An ADS that satisfies legal compliance
    but violates behavioral norms creates interaction hazards and erodes
    public trust~\cite{fridman2019}.

  \item[\propbadge{comfgreen}{Comfort}]
    Captures the rider's continuous experience of control predictability
    across three axes: longitudinal dynamics (acceleration/deceleration
    smoothness, ramp profiles), lateral dynamics (lateral jerk, steering
    abruptness, cornering feel), and onset frequency (abruptness of
    maneuver initiation). Comfort is not a luxury metric: harsh dynamics
    that are technically safe erode rider trust, reduce adoption, and
    frequently indicate edge-case behaviors warranting safety review.
    Comfort therefore serves as a proxy signal for safety margin
    monitoring.

  \item[\propbadge{effamber}{Efficiency}]
    Captures four sub-dimensions: (1)~\emph{mission completion} -
    reaching the goal without unnecessary aborts or fallbacks;
    (2)~\emph{trip time} - route efficiency, unnecessary stops;
    (3)~\emph{energy/fuel} - smooth dynamics, predictive braking;
    (4)~\emph{product-level soft metrics} - domain-specific commercial
    requirements such as bay throughput (Hub), pickup punctuality
    (robotaxi), and dwell time. Efficiency is the most underspecified
    property in academic ADS literature but is a first-class requirement
    in commercial deployment, particularly in the Hub domain.
\end{description}

\subsection{Longitudinal and Lateral Decomposition}
\label{sec:lonlat}

Every behavioral competency is further tagged by the primary control
axis it stresses: \textbf{Longitudinal (Lon)}, \textbf{Lateral (Lat)},
or \textbf{Both}. This decomposition is not merely taxonomic; it
reflects the architecture of ADS planning and control stacks, which
typically separate longitudinal and lateral controllers. Behaviors
stressing both axes simultaneously are, as a class, the hardest to
specify and validate: their longitudinal and lateral sub-specifications
interact, and failure modes in one axis can propagate to the other.
Among our 21 behaviors, all three inter-domain deep-dive examples
(Lane Change, Unprotected Left Turn, Staging Entry and Docking)
belong to the Both category.

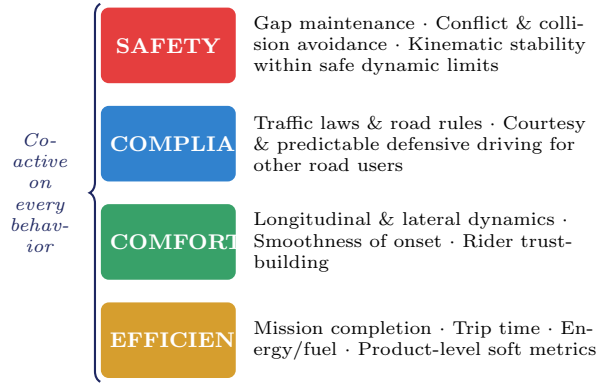
\begin{figure}[!htb]
  \centering
  \input{figures/fig_4property.tex}
  \caption{The four-property framework. Properties are co-active
           constraints on every behavioral competency; the tension
           between them defines where specification is hardest.}
  \label{fig:4prop}
\end{figure}

%% file: figures/fig_4property.tex

\begin{tikzpicture}[
  prop/.style={
    rectangle, rounded corners=3pt,
    text width=1.55cm, align=center,
    minimum height=1.0cm,
    font=\footnotesize\bfseries, text=white
  },
  desc/.style={
    rectangle, text width=4.6cm, align=left,
    font=\scriptsize, inner sep=2pt
  }
]

\node[prop, fill=safetyred] (s) at (0,0) {SAFETY};
\node[desc, right=0.15cm of s] {
  Gap maintenance $\cdot$ Conflict \& collision avoidance $\cdot$
  Kinematic stability within safe dynamic limits
};

\node[prop, fill=compblue] (c) at (0,-1.3) {COMPLIANCE};
\node[desc, right=0.15cm of c] {
  Traffic laws \& road rules $\cdot$ Courtesy \& predictable
  defensive driving for other road users
};

\node[prop, fill=comfgreen] (cf) at (0,-2.6) {COMFORT};
\node[desc, right=0.15cm of cf] {
  Longitudinal \& lateral dynamics $\cdot$
  Smoothness of onset $\cdot$ Rider trust-building
};

\node[prop, fill=effamber] (e) at (0,-3.9) {EFFICIENCY};
\node[desc, right=0.15cm of e] {
  Mission completion $\cdot$ Trip time $\cdot$
  Energy/fuel $\cdot$ Product-level soft metrics
};

\draw[decorate,
      decoration={brace, amplitude=4pt, mirror},
      thick, color=navyblue]
  (-0.9, 0.55) -- (-0.9, -4.45)
  node[midway, left=5pt, font=\scriptsize\itshape,
       text width=1.0cm, align=center, color=navyblue]
  {Co-active on every behavior};

\end{tikzpicture}

%% file: sections/04_taxonomy.tex
\section{Operational Domain Taxonomy}

\subsection{Deriving Domains from ODD Layer Composition}
\label{sec:domains}

The three operational domains in this taxonomy are not arbitrary
categorizations. They emerge from the characteristic composition of
PEGASUS ODD layers that define each deployment context:

\begin{itemize}
  \item \textbf{Highway (HWY)}: dominant layers L1 (structured lane
    geometry), L2 (traffic control infrastructure and applicable traffic laws), L4 (vehicle-only agents), L5 (high-speed conditions).
    Characterized by continuous DDT execution, structured geometry,
    and absence of complex agent interactions.
  \item \textbf{Urban (URB)}: dominant layers L1-L4 (intersections,
    mixed agents, VRUs, traffic control infrastructure, and applicable traffic laws), L5. Characterized
    by heterogeneous agent interactions, negotiated right-of-way, and
    high behavioral norm demands.
  \item \textbf{Hub (HUB)}: dominant layers L1 (facility geometry),
    L2 (facility-specific rules, not traffic law), L3 (dock events,
    bay state), L6 (digital coordination, bay assignment). Characterized
    by geofenced operation, low-speed precision, internal ruleset
    compliance, and high DDT-F concentration.
\end{itemize}

The Hub domain is structurally distinct in two important ways. First,
its compliance layer references facility-internal operational rules in addition to applicable traffic law, meaning that HWY/URB compliance specifications are not directly transferable and must be supplemented with facility-specific requirements (this taxonomic distinction does not affect or limit the application of any applicable federal, state, or local legal or regulatory requirements to hub operations). Second, it has the highest concentration of
DDT-F behaviors of any domain, yet almost no published benchmark
coverage exists for hub-specific behavioral evaluation (see
Section~\ref{sec:discussion}).

\subsection{Behavior Anatomy Template}

Each of the 21 behaviors is specified using a uniform five-field
anatomy template:

\begin{enumerate}
  \item \textbf{Trigger Condition}: the ODD state, navigation request,
    or OEDR detection event that activates the behavioral mode.
  \item \textbf{Entry/Exit Criteria}: state transitions defining when
    the ADS enters and exits the behavioral mode.
  \item \textbf{OEDR Requirements}: the object classes, events, and
    road features that must be detected and responded to.
  \item \textbf{Applicable Metrics}: Dynamic Driving Task Assessment
(DA) metrics from SAE J3237~\cite{j3237} applicable to this behavior,
selected by observability level (direct measurement, inference from
observable proxies, or internal system state access).
  \item \textbf{ODD Sensitivity}: how PEGASUS layer parameterizations
    modulate the specification (e.g., speed range, weather, agent
    density, visibility).
\end{enumerate}

Each behavior is additionally annotated with: (a) its Lon/Lat/Both
axis classification, and (b) a four-property profile indicating the
relative load on each of the Safety, Compliance, Comfort, and Efficiency
properties.

\subsection{The 21-Behavior Taxonomy}

Table~\ref{tab:taxonomy} enumerates all 21 behavioral competencies
organized by domain and Lon/Lat classification.

\begin{table}[!htb]
\fontsize{8}{10}\selectfont
\centering
\caption{21-Behavior Taxonomy: Domain, Axis Classification, and
         Primary Property Load.
         Note: this is for illustration purposes and is not meant to be comprehensive}
\label{tab:taxonomy}
\begin{tabular}{
  L{0.28\linewidth}
  C{0.08\linewidth}
  C{0.08\linewidth}
  L{0.44\linewidth}
}
\toprule
\textbf{Behavior} & \textbf{Dom.} & \textbf{Axis} & \textbf{Primary Properties} \\
\midrule
\multicolumn{4}{l}{\textit{Highway Domain (HWY)}} \\
Speed Limit Compliance           & HWY & Lon  & Compliance, Safety \\
Following Distance Maintenance   & HWY & Lon  & Safety, Efficiency \\
Emergency Braking Response       & HWY & Lon  & Safety \\
Lane Keeping                     & HWY & Lat  & Safety, Comfort \\
Cut-In Response                  & HWY & Lat  & Safety \\
Lane Change                      & HWY & Both & Safety, Compliance, Comfort \\
Highway Merge                    & HWY & Both & Safety, Compliance \\
Emergency Vehicle Response       & HWY & Both & Compliance, Safety \\
\midrule
\multicolumn{4}{l}{\textit{Urban Domain (URB)}} \\
Traffic Signal Compliance        & URB & Lon  & Compliance, Safety \\
Stop Sign Compliance             & URB & Lon  & Compliance \\
Yield / Right-of-Way             & URB & Lon  & Compliance, Safety \\
Speed Zone Compliance            & URB & Lon  & Compliance \\
Crosswalk / VRU Avoidance        & URB & Lat  & Safety \\
Protected Turn Execution         & URB & Lat  & Compliance, Safety \\
Unprotected Left Turn            & URB & Both & Safety, Compliance \\
Intersection Negotiation         & URB & Both & Safety, Compliance \\
\midrule
\multicolumn{4}{l}{\textit{Hub Domain (HUB)}} \\
Staging Queue Management         & HUB & Lon  & Efficiency, Safety \\
Docking Approach Speed Control   & HUB & Lon  & Safety, Comfort \\
Bay Alignment                    & HUB & Lat  & Safety, Efficiency \\
Geofence Boundary Compliance     & HUB & Lat  & Compliance \\
Staging Area Entry \& Docking    & HUB & Both & Safety, Efficiency \\
\bottomrule
\end{tabular}
\end{table}

\begin{figure}[!htb]
  \centering
  \includegraphics[width=\columnwidth]{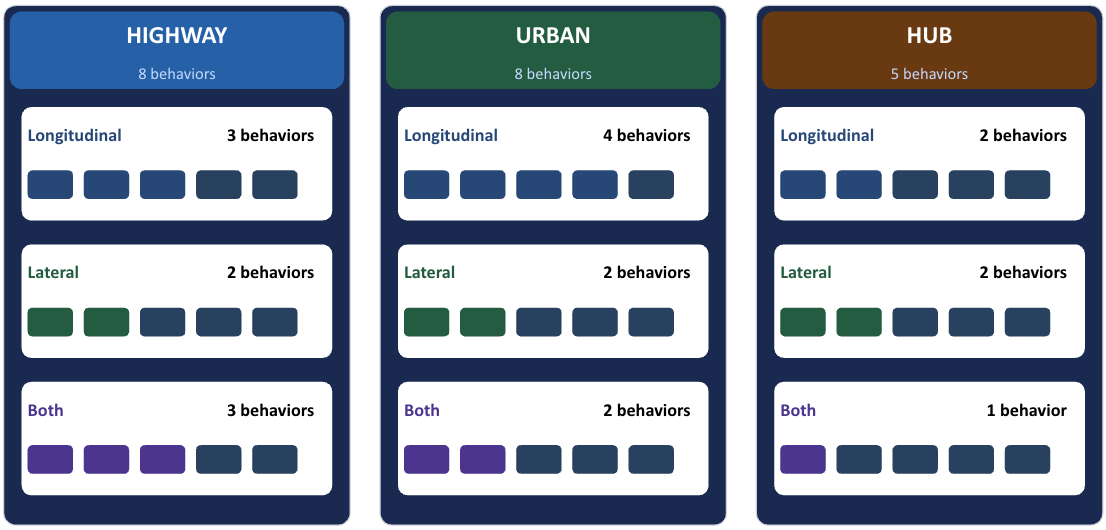}
  \caption{Taxonomy overview: 21 behaviors across
           HWY, URB, and HUB domains, with Lon/Lat/Both axis
           distribution per domain.}
  \label{fig:taxonomy}
\end{figure}

\noindent
Several structural observations merit emphasis:

\begin{itemize}
  \item The HWY domain is longitudinal-heavy (3 Lon, 2 Lat, 3 Both),
    reflecting its structured geometry and speed-dominant challenge space.
  \item The URB domain is longitudinal-heavy in regulatory behaviors
    (4 Lon) but its highest-complexity behaviors (Unprotected Left,
    Intersection Negotiation) require both axes simultaneously.
  \item The HUB domain has proportionally the highest Both-axis
    concentration (1 of 5, but the single Both behavior is its most
    operationally critical), and its compliance layer references
    facility rules rather than traffic law.
  \item Behaviors in the Both category across all domains consistently
    carry the highest Safety property load, confirming the intuition
    that multi-axis behavioral stress correlates with safety criticality.
\end{itemize}

%% file: sections/05_deepdives.tex
\section{Behavior Deep Dives}

To illustrate the behavior anatomy template in practice, this section
provides full competency specifications for one representative behavior
per domain. The three behaviors - Lane Change (HWY), Unprotected Left
Turn (URB), and Staging Area Entry and Docking (HUB) - are selected
because each is the most complex Both-axis behavior in its domain and
collectively demonstrate the full range of property loading and ODD
sensitivity across the taxonomy.

\subsection{HWY: Lane Change}
\label{sec:lc}

\noindent\textbf{Trigger}: Navigation goal or obstacle avoidance request
requiring lateral displacement to an adjacent lane.

\noindent\textbf{Entry}: Target gap in adjacent lane exceeds minimum
TTC threshold; lane change permitted by geometry and signage.

\noindent\textbf{Exit}: Ego vehicle centered in target lane, lateral
acceleration below comfort threshold; original lane clear of following
obligations.

\noindent\textbf{Lon sub-spec}: Gap acceptance against lead vehicle in
target lane ($\text{TTC} \geq \tau_{\min}$, Time Gap $\geq \Delta_{\min}$);
speed matching to target lane flow prior to merge initiation.

\noindent\textbf{Lat sub-spec}: Lane departure timing; lateral
acceleration profile during merge ($|a_y| \leq a_{y,\max}$); lateral
jerk onset; time-to-lane-center (TTLC) post-merge.

\noindent\textbf{OEDR}: Adjacent lane occupancy, TTC with approaching
vehicles, turn signal intent of neighboring agents.

\noindent\textbf{Metrics (J3237)}: Lane Departure Violation post-merge
(black-box); Aggressive Acceleration Violation (AAV) for lateral
excursions during merge (black-box); Safety Envelope Violation for
target-lane gap acceptance (black-box).

\noindent\textbf{ODD Sensitivity}: Minimum gap thresholds ($\tau_{\min}$,
$\Delta_{\min}$) increase under L5 degraded conditions (rain, night,
fog); lane width modulates acceptable lateral jerk profile; speed range
constrains available merge window.


\subsection{URB: Unprotected Left Turn}
\label{sec:ult}

Among all 21 behaviors, the Unprotected Left Turn carries the highest
aggregate Safety load. It simultaneously activates all three Safety
sub-components (gap maintenance for oncoming traffic, conflict/collision
avoidance for VRUs in crosswalks, kinematic stability through the turn
arc) and engages both Compliance layers (signal phase, right-of-way
obligation, and courtesy yielding behavior).

\noindent\textbf{Trigger}: Navigation goal requires left turn at
unprotected intersection (no dedicated left-turn phase).

\noindent\textbf{Entry}: Traffic signal permits movement; oncoming gap
accepted; no VRU in conflict crosswalk.

\noindent\textbf{Exit}: Ego vehicle centered in receiving lane;
intersection cleared; crosswalk no longer in conflict zone.

\noindent\textbf{Lon sub-spec}: Oncoming vehicle gap acceptance
($\text{TTC}_{\text{oncoming}} \geq \tau_{\text{ULT}}$); deceleration
profile to yield point; clearance timing through intersection.

\noindent\textbf{Lat sub-spec}: Turn arc geometry (minimum radius,
lane boundary constraints); crosswalk incursion avoidance; lateral
positioning in receiving lane.

\noindent\textbf{OEDR}: Oncoming vehicle TTC, VRU presence and velocity
in near and far crosswalks, traffic signal phase and countdown state,
right-of-way negotiation with opposing left-turning vehicles.

\noindent\textbf{Metrics (J3237)}: Safety Envelope Violation for
oncoming-gap conflict (black-box); Traffic Law Violation for
right-of-way/signal compliance (black-box); Crash Instance for
VRU/crosswalk conflict outcome (black-box); Event Response Time
Violation for reaction to oncoming or VRU hazards (black-box).

\noindent\textbf{ODD Sensitivity}: Oncoming gap threshold ($\tau_{\text{ULT}}$)
increases significantly under L5 low-visibility conditions; pedestrian
density (L4) modulates crosswalk dwell time; unsigned intersections
(no L2 signal) require full negotiation via behavioral norms only.


\subsection{HUB: Staging Area Entry and Docking}
\label{sec:dock}

The Staging Area Entry and Docking behavior is the most operationally
distinctive behavior in the taxonomy for three reasons. First, it is
the only behavior in which the Efficiency property load rivals Safety.
Second, its compliance layer references facility-internal operational rules in addition to applicable traffic law, meaning that HWY/URB compliance specifications are not directly transferable and must be supplemented with facility-specific requirements (this taxonomic distinction does not affect or limit the application of any applicable federal, state, or local legal or regulatory requirements to hub operations).
Third, it contains the highest DDT-to-DDT-F transition risk of any behavior: the handoff zone at bay approach is an area warranting continued engineering attention given the complexity of the DDT-to-DDT-F transition.

\noindent\textbf{Trigger}: Navigation route endpoint is a hub facility;
ego vehicle approaches facility geofence boundary.

\noindent\textbf{Entry}: Geofence crossed; speed below threshold;
assigned bay confirmed via L6 digital channel.

\noindent\textbf{Exit}: Ego vehicle at rest in assigned bay within
docking accuracy tolerance; DDT-F handoff complete.

\noindent\textbf{Lon sub-spec}: Speed ramp-down profile from facility
entry to bay approach; queue gap maintenance to vehicle ahead in staging
lane; final docking deceleration profile.

\noindent\textbf{Lat sub-spec}: Bay alignment (lateral offset from dock
marker $\leq \delta_{\max}$); geofence boundary compliance; dock marker
tracking during final approach.

\noindent\textbf{OEDR}: Bay occupancy state, ground crew presence and
clearance signal, dock alignment marker detection, queue vehicle
following distance.

\noindent\textbf{Metrics (J3237)}: [Outside the ODD] ADS DDT Execution
Violation for DDT-F handoff correctness at bay approach (grey-box);
Intervention Request/Prompt to Take Over Violation for handoff
signaling (grey-box); ODD Recognition Violation for geofence boundary
detection (white-box); DDT-Relevant Object Distance Calculation Error
Rate for dock-marker/bay-alignment accuracy (white-box).

\noindent\textbf{ODD Sensitivity}: Indoor vs.\ outdoor bay geometry
modulates sensor availability; structured vs.\ unstructured bay layout
modulates marker detection reliability; dock crew protocols vary by
facility operator (L2 layer).


%% file: sections/06_scenarios.tex
\section{ODD $\times$ Behavior $\rightarrow$ Scenario Families}

\subsection{Formal Parameterization}
\label{sec:scengen}

A key operational property of the taxonomy is that its intersection
with PEGASUS ODD layer parameterizations directly generates concrete
scenario families for behavioral testing and evaluation. We formalize
this as follows.

Let $\mathcal{B} = \{b_1, \ldots, b_{21}\}$ denote the set of 21
behavioral competencies, and let $\mathcal{L} = \{L_1, \ldots, L_6\}$
denote the PEGASUS ODD layer set. Each layer $L_i$ is associated with
a parameter space $\Omega_i$ (e.g., $\Omega_5$ includes weather state,
lighting condition, and road surface friction). A \emph{scenario family}
$\mathcal{S}(b_i, \omega)$ is defined as:

\begin{equation}
  \mathcal{S}(b_i, \omega) \;=\; b_i \;\otimes\; \omega,
  \quad \omega \in \Omega_1 \times \cdots \times \Omega_6
  \label{eq:scengen}
\end{equation}

\noindent where $\otimes$ denotes the contextualization of behavior
$b_i$ under ODD parameter instantiation $\omega$. Each $\mathcal{S}$
is associated with:

\begin{itemize}
  \item A set of entry conditions derived from the behavior's trigger
    and entry criteria (Section~4), modulated by $\omega$.
  \item A set of pass/fail acceptance criteria derived from the
    behavior's applicable J3237 metrics, with thresholds modulated
    by $\omega$ (e.g., TTC thresholds increase under low visibility).
  \item An ODD sensitivity profile indicating which layers have the
    strongest modulating effect on the scenario's difficulty and
    metric thresholds.
\end{itemize}

\subsection{Worked Example}

\noindent\textbf{Behavior}: Lane Change (HWY, Section~\ref{sec:lc})

\noindent\textbf{ODD instantiation $\omega$}:
\begin{itemize}
  \item $L_1$: three-lane divided highway, \SI{3.6}{\meter} lanes
  \item $L_4$: cut-in agent in target lane, initial TTC = 1.5\,s
  \item $L_5$: rain, nighttime, road surface friction coefficient 0.6
\end{itemize}

\noindent\textbf{Scenario family $\mathcal{S}$}:
\begin{itemize}
  \item Entry: ego navigating in left lane; obstacle in path triggers
    lane change request to center lane; cut-in agent approaches at
    closing speed.
\item Pass criteria: $\text{TTC}_{\text{adj}} \geq \tau_{\min}(\omega)$
    where $\tau_{\min}$ is increased under L5 rain/night conditions
    per ODD sensitivity profile (an illustrative 20\% shown here for
    concreteness; actual factors are ODD and implementation-specific);
    $|a_y| \leq a_{y,\max}$
    at road friction coefficient; lateral deviation from lane center
    $\leq \delta_{\text{TTLC}}$ within \SI{3}{\second} of completion.
  \item Fail modes: merge initiated below minimum gap; lateral jerk
    exceeds comfort envelope; lane center not achieved within TTLC
    threshold.
\end{itemize}

This single behavior $\times$ ODD instantiation yields a concrete,
executable scenario with defined pass/fail criteria traceable to J3237
metrics (Safety Envelope Violation, AAV) and supplementary
comfort-oriented criteria (TTLC), modulated by PEGASUS layer
parameters.

\subsection{SOTIF Alignment}

The scenario family generation mechanism is designed to support the
ISO~21448 evidence structure~\cite{sotif2022}. SOTIF describes identification of triggering conditions for known unsafe
scenarios; the ODD sensitivity profiles in each behavior specification
identify which layer parameters most strongly modulate behavioral stress,
intended to inform triggering condition analysis. SOTIF also describes verification and validation based on scenarios; scenario families
generated via Equation~\ref{eq:scengen} are suitable for use within a SOTIF verification and validation, with the behavioral competency coverage metric
potentially serving as one input to a coverage argument.

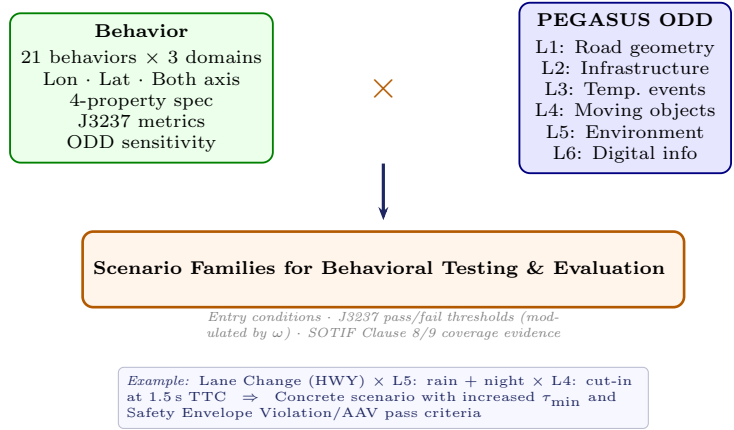
\begin{figure}[!htb]
  \centering
  \input{figures/fig_scenarios.tex}
  \caption{ODD $\times$ Behavior $\rightarrow$ Scenario Families.
           Each PEGASUS layer parameterization crossed with a behavior specifications yields a concrete scenario family
           with J3237-traceable pass/fail criteria.}
  \label{fig:scenarios}
\end{figure}

\subsection{Fleet Log Coverage Verification}

An important complementary use of the taxonomy is as a coverage
query schema for AV fleet log mining. Given a fleet log corpus and
the 21-behavior taxonomy, each log segment can be labeled against
the behavior it exercises and the ODD parameters under which it
was collected, enabling computation of behavioral competency coverage
scores across the corpus. This application is a direct extension of
the scenario family generation mechanism and is identified as future
work in Section~\ref{sec:future}.

%% file: figures/fig_scenarios.tex

\begin{tikzpicture}[
  box/.style={
    rectangle, rounded corners=4pt,
    minimum width=2.8cm, minimum height=1.6cm,
    draw, line width=0.7pt,
    font=\scriptsize, align=center, inner sep=4pt
  },
  outbox/.style={
    rectangle, rounded corners=4pt,
    minimum width=6.8cm, minimum height=1.0cm,
    draw, line width=1.0pt,
    font=\scriptsize\bfseries, align=center, inner sep=4pt
  },
  arr/.style={-{Stealth[length=5pt]}, thick}
]

\node[box, fill=blue!10, draw=blue!50!black] (odd) at (6.4,0) {
  \textbf{PEGASUS ODD}\\[2pt]
  \scriptsize
  L1: Road geometry\\
  L2: Infrastructure\\
  L3: Temp.\ events\\
  L4: Moving objects\\
  L5: Environment\\
  L6: Digital info
};

\node[font=\Large\bfseries, text=orange!70!black] (times) at (3.2,0) {$\times$};

\node[box, fill=green!8, draw=green!50!black] (beh) at (0,0) {
  \textbf{Behavior}\\[2pt]
  \scriptsize
  21 behaviors $\times$ 3 domains\\
  Lon $\cdot$ Lat $\cdot$ Both axis\\
  4-property spec\\
  J3237 metrics\\
  ODD sensitivity
};

\draw[arr, line width=1.2pt, color=navyblue] (3.2,-1.0) -- (3.2,-1.75);

\node[outbox, fill=orange!8, draw=orange!70!black] (out) at (3.2,-2.4) {
  Scenario Families for Behavioral Testing \& Evaluation
};

\node[font=\tiny\itshape, text=gray, text width=7cm, align=center]
  at (3.2,-3.15) {%
  Entry conditions $\cdot$ J3237 pass/fail thresholds (modulated by $\omega$)
  $\cdot$ SOTIF Clause~8/9 coverage evidence};

\node[font=\tiny, text=navyblue, text width=6.8cm, align=left,
      draw=navyblue!30, rounded corners=2pt, fill=blue!3,
      inner sep=3pt] at (3.2,-4.1) {%
  \textit{Example:} Lane Change (HWY) $\times$
  L5: rain + night $\times$ L4: cut-in at 1.5\,s TTC
  $\;\Rightarrow\;$ Concrete scenario with increased $\tau_{\min}$
  and Safety Envelope Violation/AAV pass criteria};

\end{tikzpicture}

%% file: sections/07_discussion.tex
\section{Discussion}
\label{sec:discussion}

\subsection{The DDT / DDT-F Dimension}

The taxonomy reveals a structural dimension that is largely absent from
existing behavioral frameworks: the DDT/DDT-F distribution across
domains. Behaviors in which the ADS maintains full continuous Dynamic
Driving Task execution (DDT behaviors) are well-represented in
existing benchmarks and simulation environments. Behaviors involving
DDT Fallback (DDT-F) - transitions, handoffs, minimal risk condition
activation - are poorly covered.

The three domains differ substantially in their DDT/DDT-F concentration.
The Highway domain is DDT-dominant: the ADS is in continuous control
throughout all eight behaviors. The Urban domain is mixed: most behaviors
are DDT, but intersection negotiation and emergency scenarios involve
partial DDT-F elements. The Hub domain is DDT-F heavy: the Staging Area
Entry and Docking behavior involves a mandatory DDT-to-DDT-F transition
at bay approach, and this transition represents the highest-complexity operational handoff in the domain. To our knowledge, no published benchmark specifically targets hub DDT-F transitions.

\subsection{The Hub Domain Gap}
\label{sec:hubgap}

The Hub domain represents the most structurally distinct and least
researched operational context in the taxonomy. Its distinguishing
characteristics - geofenced operation, facility-rule-based compliance,
low-speed precision maneuvering, high DDT-F concentration, and
Efficiency as a first-class property - mean that HWY and URB
behavioral specifications are not directly transferable. Yet the
published literature treating hub-specific behavioral requirements remains limited. Public datasets and benchmarks for hub behavioral evaluation are not currently available at the scale of those for highway and urban domains
(e.g., Argoverse~\cite{argoverse2}, Waymo Open Motion~\cite{womd}).

This gap is consequential: commercial AV deployments increasingly
rely on hub operations as the operational anchor for service
(depot-to-service-area dispatch, automated last-mile logistics). The
five Hub behaviors in this taxonomy, and particularly Staging Area
Entry and Docking, represent a high-priority research target.

\subsection{Open Problems in the Published Literature}

The taxonomy surfaces five gaps in the published academic and industry
literature on ADS behavioral specification and validation.

\textit{Note that these open problems describe gaps in the published
research literature and are not intended to characterize the completeness
or adequacy of any specific organization's internal validation or safety
assurance processes, which may address these areas through methods not
described in the public literature.}

\begin{enumerate}
  \item \textbf{Inter-behavior transition specifications}: Formal
    specification of behavioral state transitions at domain boundaries
    (e.g., HWY to URB entry) and between behavioral modes within a
    domain does not appear in the published literature. The behavior
    anatomy template specifies entry and exit criteria per behavior
    but does not formalize the transition graph between behaviors.

  \item \textbf{Quantitative competency coverage metrics}: To our
    knowledge, no publicly documented metric currently exists for
    measuring the degree to which a scenario suite covers a behavioral
    specification space. A Behavioral Competency Coverage Score (BCCS)
    over a taxonomy such as the one presented here would provide a
    tractable, standards-aligned coverage argument suitable for public
    disclosure.

  \item \textbf{Hub domain benchmarks}: The development of publicly
    available behavioral benchmarks and annotated datasets for hub
    operational behaviors represents a gap in the published literature
    with direct deployment relevance.

  \item \textbf{Closed-loop behavioral compliance at scale}: Open-loop
    metric evaluation does not capture interaction dynamics between ADS
    and other agents. To our knowledge, no publicly documented
    infrastructure exists for behavior-level compliance evaluation in
    reactive multi-agent scenarios at deployment scale.

  \item \textbf{Four-property weighting by ODD context}: The relative
    weighting of Safety, Compliance, Comfort, and Efficiency properties
    is likely ODD-dependent. To our knowledge, no publicly documented
    framework for context-dependent property weighting in multi-objective
    behavioral specification has been proposed in the literature.
\end{enumerate}

%% file: sections/08_conclusions.tex
\section{Summary and Conclusions}

This paper presented a structured, standards-grounded taxonomy of 21
behavioral competencies for automated driving systems, organized across
three operationally distinct domains derived from PEGASUS ODD layer
composition. The primary contributions are:

\begin{enumerate}
  \item \textbf{A derivation methodology}: Operational domains and
    behavioral competencies are derived systematically from ODD layer
    composition, providing a principled bridge from operating condition
    specification to behavioral enumeration.

  \item \textbf{A 21-behavior taxonomy}: Behaviors are specified using a
    uniform five-field anatomy template, decomposed along longitudinal
    and lateral control axes, and characterized against a four-property
    framework (Safety, Compliance, Comfort, Efficiency) that is
    simultaneously active on every behavior.

  \item \textbf{A scenario generation mechanism}: The formal crossing
    of PEGASUS ODD layer parameterizations with behavioral competency
    specifications yields concrete scenario families with J3237-traceable
    pass/fail criteria, directly supporting SOTIF Clause~8/9 coverage
    evidence generation.
\end{enumerate}

The taxonomy identifies the Hub domain as the most structurally
distinct and least benchmarked operational context in commercial ADS
deployment, with the DDT-F transition in Staging Area Entry and
Docking representing a high-priority gap for both specification and
validation research.

\subsection{Future Research Directions}
\label{sec:future}

This taxonomy serves as the behavioral specification layer for a
broader research program. The following directions are identified
as extensions warranted by gaps in the published literature, and
are not intended to suggest that these capabilities are absent from
any organization's internal engineering or validation processes.

Immediate directions include: (1) formal specification of the
inter-behavior transition graph at domain boundaries; (2) development
and public documentation of a Behavioral Competency Coverage Score
(BCCS) metric for scenario suite adequacy assessment, of which to
our knowledge no publicly documented version currently exists;
(3) publication of fleet log mining methodology for behavioral
coverage verification using the taxonomy as a query schema;
(4) closed-loop behavioral compliance evaluation in reactive
simulation environments; and (5) hub-specific benchmark
development for public release. The taxonomy has been deployed
as the specification layer for a rule-enforced trajectory
optimization system~\cite{rector2026}, establishing its
operational validity as an engineering artifact.

\vspace{6pt}
\noindent\small\textit{For the avoidance of doubt, this publication
is not intended and shall not be construed to establish, create, or
be deemed to have created a required or expected standard, methodology,
or benchmark regarding the design, testing, or validation of automated
driving systems, nor shall it prevent, hinder, or restrict Torc
Robotics, Inc.\ or any affiliate from adopting, rejecting, or
deviating from any concept, framework, or open problem discussed
herein in its actual engineering practices. References to open
research problems, future work, or gaps in published literature
describe the state of the broader academic and industry research
landscape and are not representations regarding the completeness,
adequacy, or sufficiency of Torc's internal validation, testing, or
safety assurance processes for any specific deployed product.}
\normalsize
\vspace{6pt}